\documentclass{llncs}
\usepackage{makeidx}  
\usepackage{amsmath} 
\usepackage{amssymb} 
\usepackage[linesnumbered,lined]{algorithm2e}
\usepackage[colorinlistoftodos]{todonotes} 
\usepackage{cite}
\usepackage{graphicx}
\usepackage{relsize}
\graphicspath{ {./Imgs/} }
\usepackage{array}
\usepackage{multirow}
\usepackage{rotating}
\usepackage{wasysym}
\usepackage{mathtools}
\usepackage{tabularx}
\usepackage{cleveref}
\usepackage{floatrow}
\usepackage[misc]{ifsym}
\begin{document}
\title{Elastic Registration of Geodesic Vascular Graphs}
\author{Stefano Moriconi\inst{1}\textsuperscript{(\Letter)} \and Maria A. Zuluaga\inst{2} \and H. Rolf J{\"a}ger\inst{3} \and Parashkev Nachev\inst{3} \and S\'ebastien Ourselin\inst{4} \and M. Jorge Cardoso\inst{4,1} }
\institute{Translational Imaging Group, CMIC, University College London
\email{stefano.moriconi.15@ucl.ac.uk}
\and
Universidad Nacional de Colombia, Bogot\'{a}, Colombia
\and
Institute of Neurology, University College London
\and
School of Biomedical Engineering and Imaging Sciences, King's College London}
\maketitle
\begin{abstract}
Vascular graphs can embed a number of high-level features, from morphological parameters, to functional biomarkers, and represent an invaluable tool for longitudinal and cross-sectional clinical inference. This, however, is only feasible when graphs are co-registered together, allowing coherent multiple comparisons. The robust registration of vascular topologies stands therefore as key enabling technology for group-wise analyses. In this work, we present an end-to-end vascular graph registration approach, that aligns networks with non-linear geometries and topological deformations, by introducing a novel over-connected geodesic vascular graph formulation, and without enforcing any anatomical prior constraint. The 3D elastic graph registration is then performed with state-of-the-art graph matching methods used in computer vision. Promising results of vascular matching are found using graphs from synthetic and real angiographies. Observations and future designs are discussed towards potential clinical applications.
\end{abstract}
\vspace{-10pt}
\section{Introduction}
\vspace{-6pt}
Vascular graphs can be obtained from angiographies using connectivity paradigms and network extraction algorithms by embedding high-level features, such as spatial location, direction, scale, and bifurcations. However, the correct extraction of subject-specific vascular topologies, in complex (cerebro)vascular networks, can be challenging when rather tortuous and tangled structures are present. In other cases, anatomical cycles and their variants (i.e. the circle of Willis, anastomoses and fenestrations) \cite{jinkins2000atlas}, the presence of pathology (e.g. tangled arterio-venous malformations, neoplastic and embryologic plexiforms), and image-related limitations (e.g. unresolved kissing vessels) dramatically increase the network complexity, and sometimes impede the extraction of the vascular topology as a tree. A viable approach is to consider a data-driven vectorial prior from an early group-wise vascular graph registration. Defining a group-wise vectorial prior first embeds the likelihood of connectivity patterns from a population, and subsequently injects a probabilistic prior towards the inference of the most meaningful subject-specific vascular topology. The same vectorial prior could also embed morphometric parameters, functional and hemodynamic descriptors and surrogate biomarkers, constituting thus a labelled multi-spectral vascular atlas. By registering the obtained vectorial atlas over a set of similar vascular graphs, a number of group-level clinical analyses would be allowed, from inter-subject comparisons of the underlying vascular morphology, to longitudinal studies of vascular pathologies, on which clinical prediction and therapeutic inference ultimately depend. The robust alignment of multiple topologies is of critical relevance and represents a methodological bottleneck for population-level analyses. The alignment of networks and vectorial graphs raised increasing interest among the scientific community in the last decade. Motivated by registering acyclically connected structures from biomedical imaging, (e.g. vascular and respiratory trees), \cite{feragen2012hierarchical, charnoz2005design, xue2006automatic, wang2017automatic, petersen2013airways, serradell2015nonrigid} introduced different registration techniques, which mostly rely on pairwise matching distances between junction nodes and connecting edges. Following an initial alignment, these methods usually minimise a similarity cost function or maximise a probabilistic likelihood between pairs of nodes/edges or sub-trees and graph kernels, and hierarchically evaluate the correspondences at different levels of tree-depth. Whilst only few formulations would register generic spatial graphs \cite{serradell2015nonrigid}, in all cases the considered topologies were either hierarchically pre-defined as trees, or determined beforehand on a specific anatomical compartment. Also, since these methods exploit node locations, branches geometry, arborescence depth, or the parent-child relation of a rooted tree, they require the explicit tree topology to accurately capture the underlying vasculature, where each bifurcation is correctly annotated as its connectivity pattern. The registration of noisy topologies, (i.e. mis-connections, missing branches and short-cuts), and non-linearly deformed geometries remains a challenging and open problem.
\begin{figure}[t!]
\label{Fig01}
\tiny
\begin{tabular}{cc|ccc}
\multicolumn{2}{c}{\textbf{Geodesic Vascular Graph}}&\multicolumn{3}{c}{\textbf{Graph Matching Problem}}\\
\includegraphics[width=0.167\textwidth]{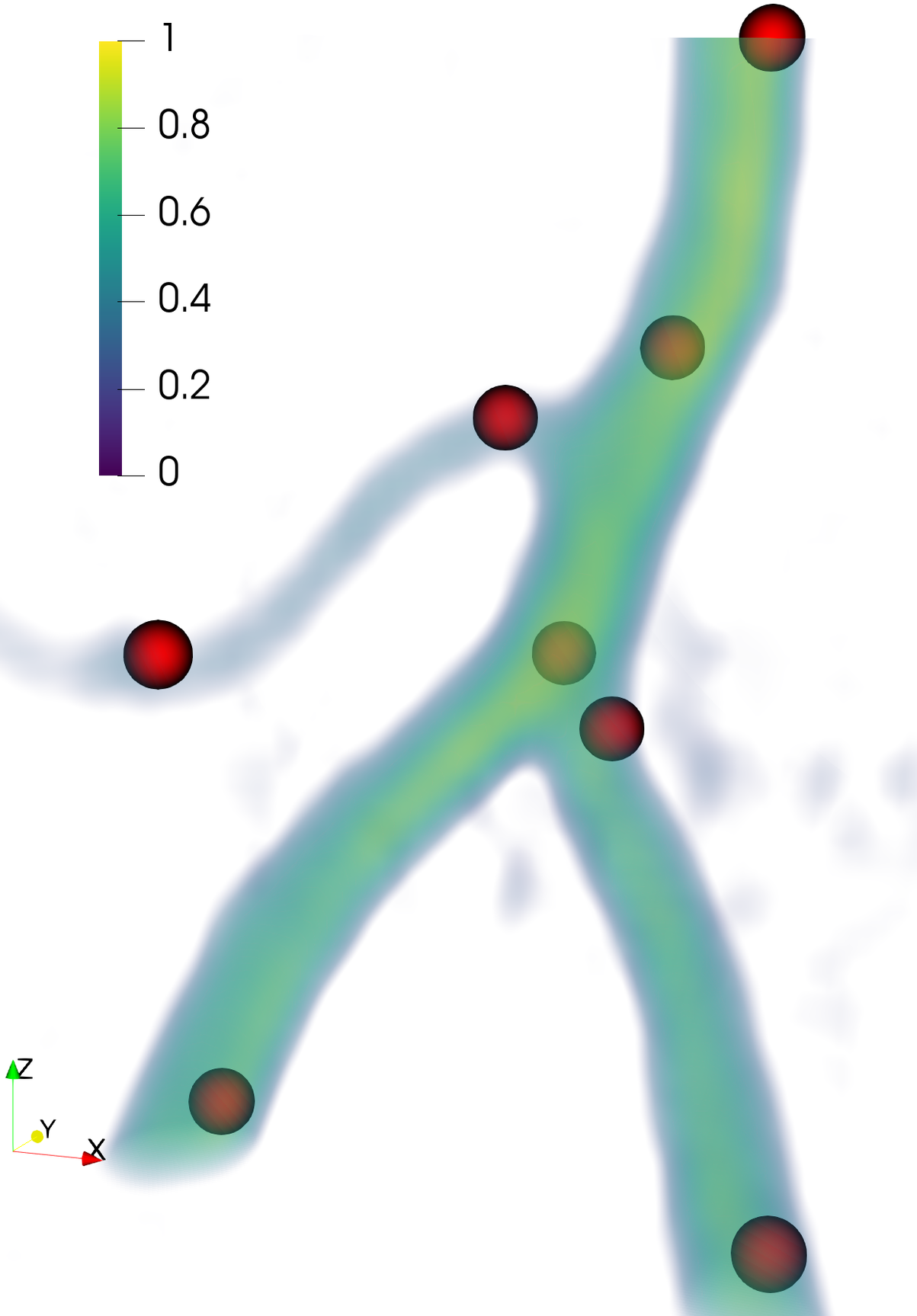}&\includegraphics[width=0.167\textwidth]{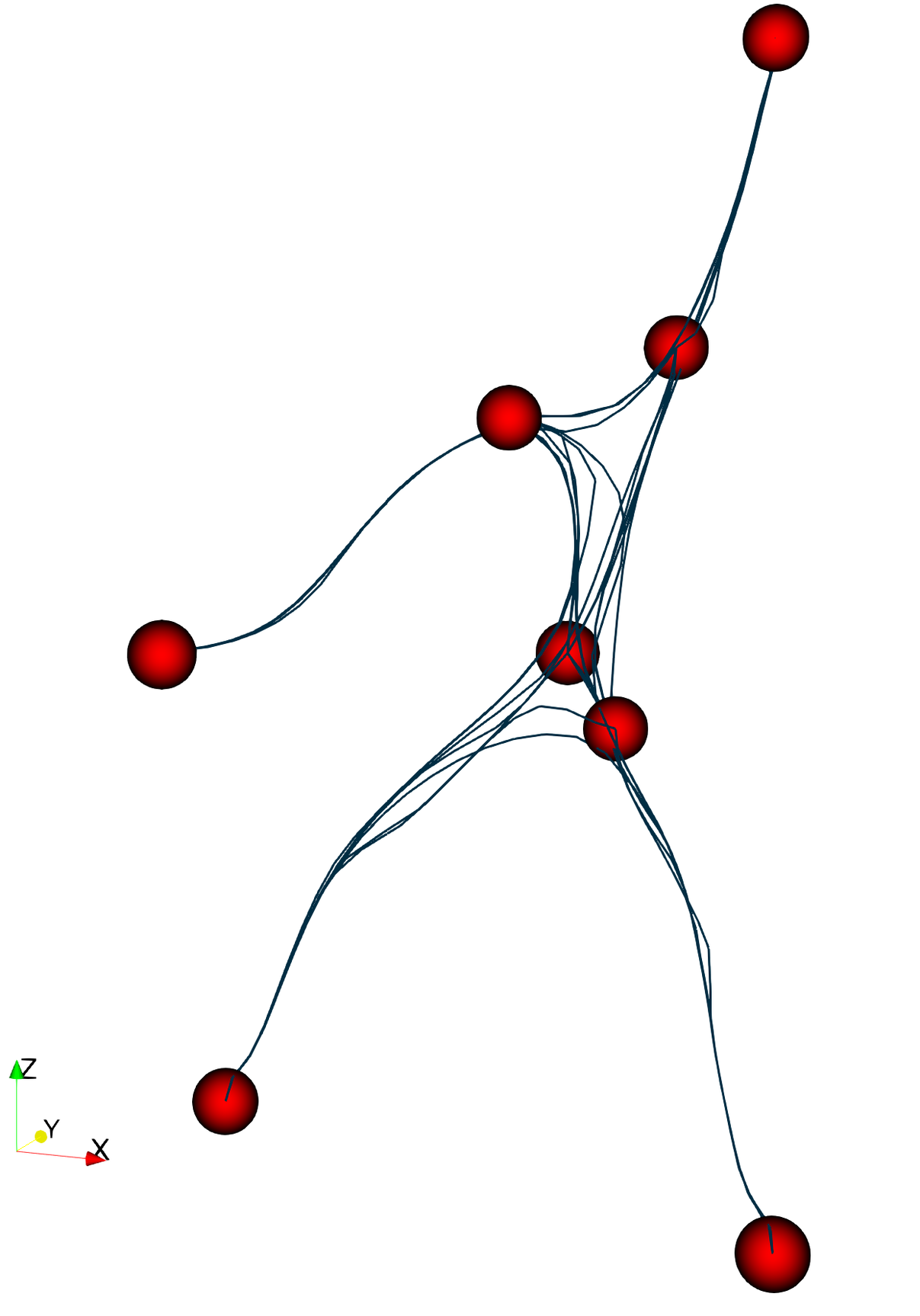}&\includegraphics[width=0.105\textwidth]{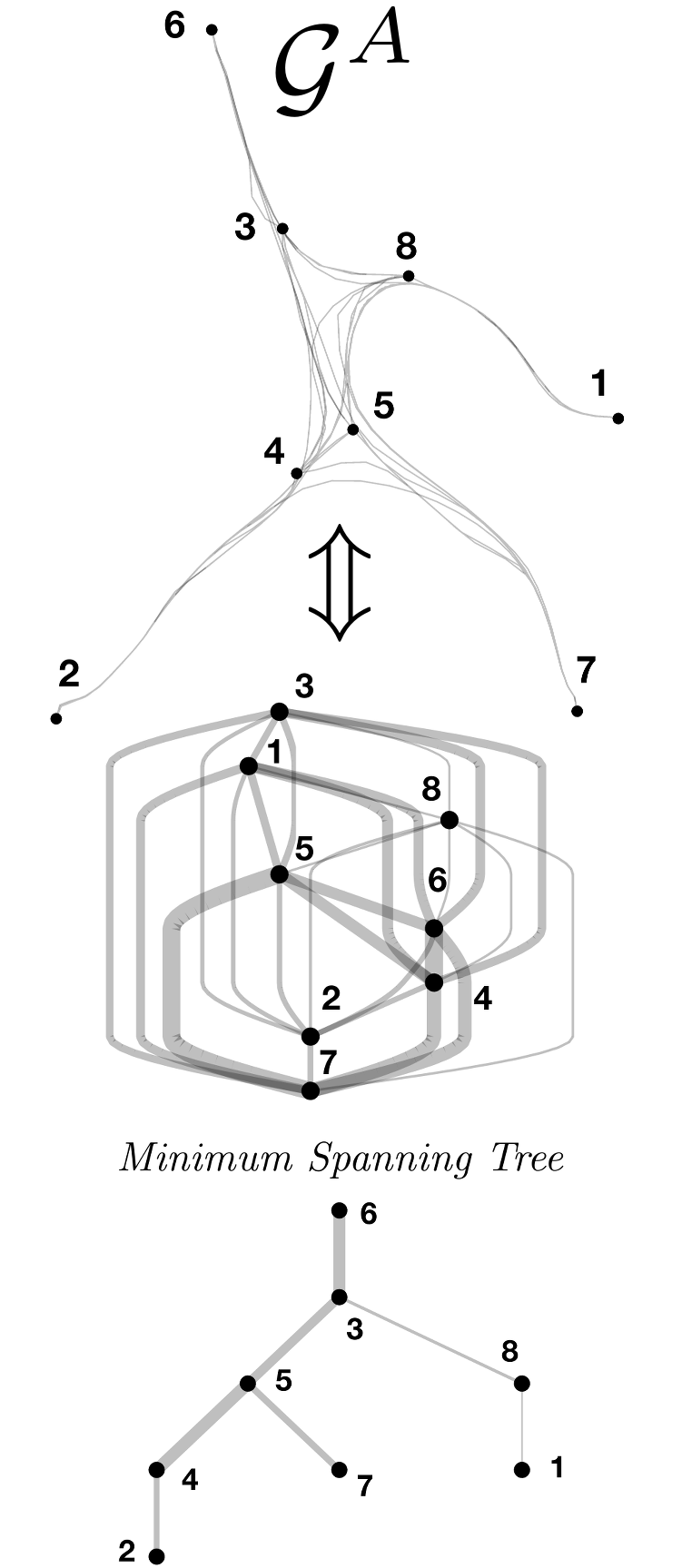}&\includegraphics[width=0.105\textwidth]{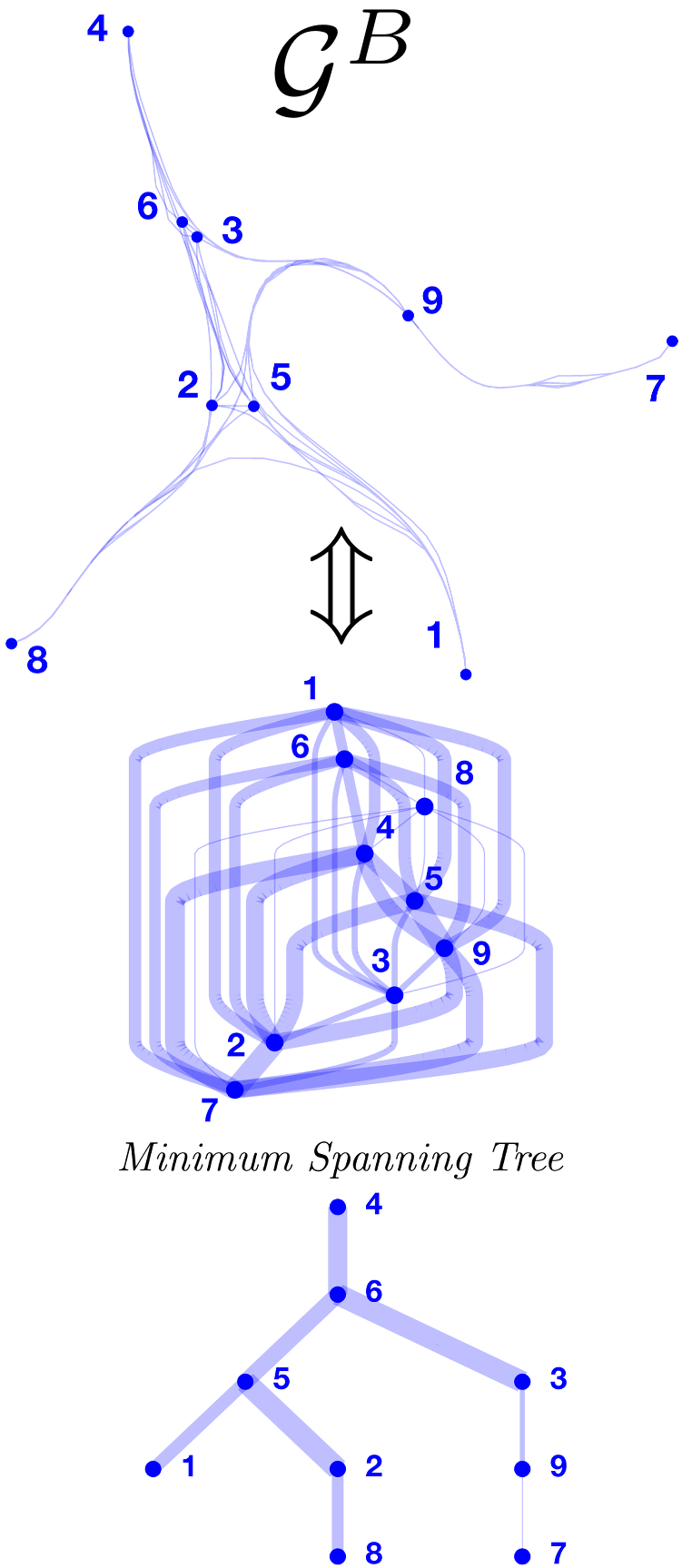}&\includegraphics[width=0.375\textwidth]{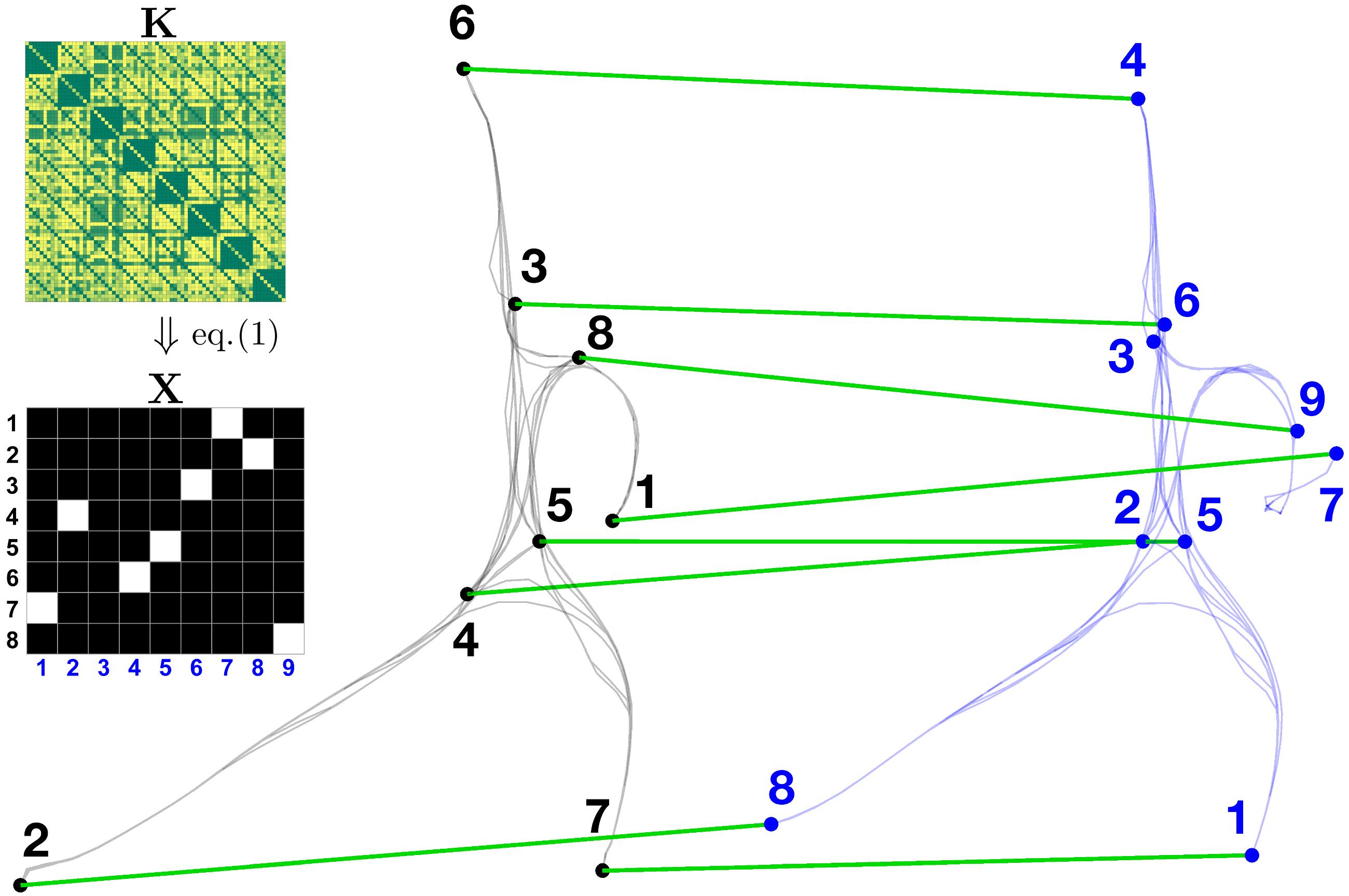}
\end{tabular}
\vskip -6pt 
\caption{Geodesic Vascular Graph and GM problem of non-linearly deformed topologies. Extraction of a fully-connected topology from an initial set of nodes (left). Associated graph representations and minimum spanning trees for two topologically different instances ($\mathcal{G}^{A}$ and $\mathcal{G}^{B}$) of the same underlying vascular anatomy (center). Graphs alignment and nodes matching for the generalised GM problem (right).}
\normalsize
\end{figure}
In this preliminary work, we address vascular graph matching (GM) by relaxing assumptions on the acyclic (un)directed graph structure and the anatomical hierarchical prior from any vascular compartment. The idea is to consider and register the vasculature as an over-connected graph: a redundant topology encoding the likelihood of connections between neighbouring nodes with minimal paths. This enhanced connectivity pattern would compensate for topological inaccuracies, for non-linear deformations of branches, and would enrich the registration space-search with distinctive features. The pairwise graph registration problem can be subsequently solved using generic GM algorithms. In the following sections, the proposed approach is first described, then, an experimental set-up is presented, comprising graphs from synthetic and real angiographies. The accuracy of different GM algorithms is evaluated on correct nodes correspondences. Observations and conclusions are discussed, focusing on future developments and potential applications.
\vspace{-10pt}
\section{Methods}
\label{section_Methods}
\vspace{-6pt}
Aiming at the pairwise alignment of vascular topologies within a deformable and anatomical prior-free framework, we first introduce a novel over-connected geodesic vascular graph (GVG), then the generic GM problem is presented together with the proposed affinity metrics based on vessels geometry and their redundant geodesic connectivity. The two-steps registration pipeline is described also listing the considered GM algorithms.
\vspace{-6pt}
\subsubsection{Geodesic Vascular Graph.}
\label{subsection_GVG}
We define the undirected geodesic vascular graph $\mathcal{G} = (N,E)$ in $\mathbb{R}^3$, as the set of nodes $\mathbf{n}_i \in N$, and the associated set of connecting geodesic edges $\mathbf{e}_v \in E$, encoding the graph adjacency list. Each geodesic edge $\mathbf{e}_v$ is defined as the 3D shortest path joining a generic pair of nodes, by solving the Eikonal equation \cite{kimmel1998computing} over a vascular smoothly connected manifold as in \cite{moriconi2017vtrails}. However, an exhaustive search is here performed by connecting all pairs of nodes independently, or up to a pre-defined spatial neighborhood $\nu$. This determines an over-connected vascular graph of minimal paths, which fully captures the underlying vasculature with enhanced geodesic redundancy \mbox{(Fig. 1)}. Together with the formulation of the over-connected $\mathcal{G}$, we also introduce a set of edge- and node-attributes. The edge-attributes $\mathbf{e}_v=\{\mathbf{p}_v , l_v , u_v\}$ comprise the dense sampling $\mathbf{p}_v$ of each shortest path in 3D (i.e. the point coordinates sequence as in Fig. 2,3), its associated euclidean length $l_v$ and the geodesic integral energy $u_v$ integrated along the path, as in \cite{moriconi2017vtrails}. The node-attributes $\mathbf{n}_i = \{\mathbf{c}_i , d_i\}$ include the spatial location $\mathbf{c}_i$ as coordinates in $\mathbb{R}^3$, and the geodesic node degree $\mathsmaller{d_i = \frac{1}{| \tilde{\mathbf{e}}_v |} \sum_{\tilde{\mathbf{e}}_v}{u_v} }$, with $\tilde{\mathbf{e}}_v$ the set of incident edges of cardinality $| \tilde{\mathbf{e}}_v |$.
\vspace{-6pt}
\subsubsection{Graph Matching Problem and Affinity Metrics.}
\label{subsection_GMp_AffMetrics}
As presented in \cite{zhou2016factorized}, the problem of matching a pair of graphs $\mathcal{G}^{A}$ and $\mathcal{G}^{B}$ requires the definition of an affinity matrix $\mathbf{K}$ to measure the similarity between each pair of nodes and edges. Given the node cardinality $i = |\mathbf{n}^{A}_{i}|$, and $j = |\mathbf{n}^{B}_{j}|$, the symmetric affinity matrix $\mathbf{K} \in \mathbb{R}^{ij \times ij}$ encodes the similarity between nodes along its diagonal elements, whereas the edges similarity is encoded in the off-diagonal ones. Given $\mathbf{K}$, the problem of graph matching consists in finding the optimal correspondence $\mathbf{X}$ between all the nodes \mbox{(Fig. 1)}, so that a compatibility functional $J(\mathbf{X})$ is maximised with a quadratic assignment problem (QAP) \cite{loiola2007survey},
\vspace{-2pt}
\begin{equation}
\label{eq1}
\max{~J(\mathbf{X})} = \text{vec}(\mathbf{X})^{t} ~ \mathbf{K} ~ \text{vec}(\mathbf{X}),
\vspace{-2pt}
\end{equation}
where $\mathbf{X}$ is constrained to be a one-to-one mapping between the sets of nodes $\mathbf{n}^{A}_{i}$ and $\mathbf{n}^{B}_{i}$, and $\text{vec}(\mathbf{X})$ denotes the vectorisation of the correspondence matrix. We formulate both node- and edge-similarity metrics for the definition of the affinity matrix $\mathbf{K}$, by adopting the matrix factorisation as in \cite{zhou2016factorized}, i.e. $\mathbf{K}_{\mathbf{n}^{\text{\textit{AB}}}}$ and $\mathbf{K}_{\mathbf{e}^{\text{\textit{AB}}}}$ respectively. In detail, we define
\begin{gather}
\label{eq2-3}
\vspace{-6pt}
\mathbf{K}_{\mathbf{n}^{\text{\textit{AB}}}} = e^{-\left( \alpha_{1} \frac{\mathbf{C}^{\text{\textit{AB}}}}{\sigma_{\mathbf{C}}} ~+~ \alpha_{2} \frac{\mathbf{D}^{\text{\textit{AB}}}}{\sigma_{\mathbf{D}}} \right)} ~~~~ \text{with} ~~ \alpha_{1} + \alpha_{2} = 1,~~~ \text{and}\\
\mathbf{K}_{\mathbf{e}^{\text{\textit{AB}}}} = e^{-\left( \beta_{1} \frac{\mathbf{P}^{\text{\textit{AB}}}}{\sigma_{\mathbf{P}}} ~+~ \beta_{2} \frac{\mathbf{L}^{\text{\textit{AB}}}}{\sigma_{\mathbf{L}}} ~+~ \beta_{3} \frac{\mathbf{U}^{\text{\textit{AB}}}}{\sigma_{\mathbf{U}}} \right)} ~~~~ \text{with} ~~ \beta_{1} + \beta_{2} + \beta_{3} = 1,
\vspace{-6pt}
\end{gather}
where $\mathbf{C}^{\text{\textit{AB}}}$ and $\mathbf{D}^{\text{\textit{AB}}}$ are the pairwise $\ell^{2}$-norm matrices between the two sets of node coordinates $\{\mathbf{c}^{A}_i,\mathbf{c}^{B}_j\}$, and geodesic degrees $\{d^{A}_i,d^{B}_j\}$, as well as $\mathbf{P}^{\text{\textit{AB}}}$, $\mathbf{L}^{\text{\textit{AB}}}$ and $\mathbf{U}^{\text{\textit{AB}}}$ are the pairwise average symmetric distance matrices of the connecting minimal paths $\{\mathbf{p}^{A}_v,\mathbf{p}^{B}_w\}$, and the pairwise $\ell^{2}$-norm matrices between the sets of the euclidean lengths $\{l^{A}_v,l^{B}_w\}$ and geodesic integral energies $\{u^{A}_v,u^{B}_w\}$, respectively.
The normalisation factors $\sigma_{\mathbf{C},\mathbf{D},\mathbf{P},\mathbf{L},\mathbf{U}}$ are the standard deviations estimated from the off-diagonal elements of the associated distance matrices over the considered population of graphs.
Lastly, $\alpha_{1}$, $\beta_{1}$ and $\beta_{2}$ weight the geometrical similarities among nodes and edges, whereas $\alpha_{2}$ and $\beta_{3}$ represent the respective geodesic trade-off.
We refer to \cite{zhou2016factorized} for the composition of $\mathbf{K}$ from the factorised components $\mathbf{K}_{\mathbf{n}^{\text{\textit{AB}}}}$ and $\mathbf{K}_{\mathbf{e}^{\text{\textit{AB}}}}$, and for the QAP solver implementation.
\vspace{-6pt}
\subsubsection{Graph Registration.}
\label{subsection_GraphRegPipeline}
Although some GM algorithms do not require any spatial initialisation of the graphs, we present a two-steps approach \mbox{(Fig. 2, 3)} by combining an early coarse alignment strategy to facilitate the further registration by reducing biases due to pure rigid mis-alignment.
\vspace{-6pt}
\paragraph{Rigid Alignment.}
The globally-optimal iterative closest point \mbox{(Go-ICP)} \cite{yang2016go} is run on $\mathcal{G}^{A}$ and $\mathcal{G}^{B}$ as coarse geometrical initialisation. Here, the dense cloud of samples, i.e. the nodes coordinates $\{\mathbf{c}^{A}_{i},\mathbf{c}^{B}_{j}\}$ and the sequences of edge points $\{\mathbf{p}^{A}_{v},\mathbf{p}^{B}_{w}\}$, is retrieved for the spatial rigid pre-alignment. \mbox{Go-ICP} searches the entire 3D motion space, and, under the minimisation of an $L_{2}$ error metric based on a branch-and-bound scheme, guarantees the global optimality of the rigid mapping, even in presence of noisy data, outliers, and partial samples overlap.
\vspace{-6pt}
\paragraph{Fine Graph Matching.}
Classic GM algorithms employed in computer vision, are considered for the fine registration. We account for Graduated Assignment (GA) \cite{gold1996graduated}, Spectral Matching (SM) \cite{leordeanu2005spectral}, Spectral Matching with Affine Constraints (SMAC) \cite{cour2007balanced}, Probabilistic Matching (PM) \cite{zass2008probabilistic}, Integer Projected Fixed Point (IPFP-U/SM) \cite{leordeanu2009integer}, Re-weighted Random Walk Matching (RRWM) \cite{cho2010reweighted}, and the current state-of-the-art, the non-rigid Factorized Graph Matching (FGM) \cite{zhou2016factorized}. The deformable graph matching problem, detailed in \cite{zhou2016factorized}, formulates the unknown graph correspondence being constrained with a geometric transformation $T$. A composition of transformations (i.e. similar, affine, and non-rigid) are incorporated into the compatibility function (\cref{eq1}), and subsequently estimated by optimising jointly the correspondence matrix $\mathbf{X}$ and the composite transformation $T$ itself. We employed the undirected-graph versions of the listed algorithms. Implementations and configurations are available from authors' websites.
\vspace{-14pt}
\section{Experiments and Results}
\label{section_ExpRes}
\vspace{-6pt}
\begin{figure}[t]
\label{Fig02}
\tiny
\begin{tabular}{cc|c|cc}
$\mathbf{\mathcal{G}^{A}}$&$\mathbf{\mathcal{G}^{A}_\mathsmaller{{\mathcal{D}_{40\%}\mathcal{T}_{30\%}}}}$&\textbf{Rigid Alignment}&\multicolumn{2}{c}{\textbf{Fine GM (FGM) - acc: 87.34$\%$}}\\
\includegraphics[width=0.195\textwidth]{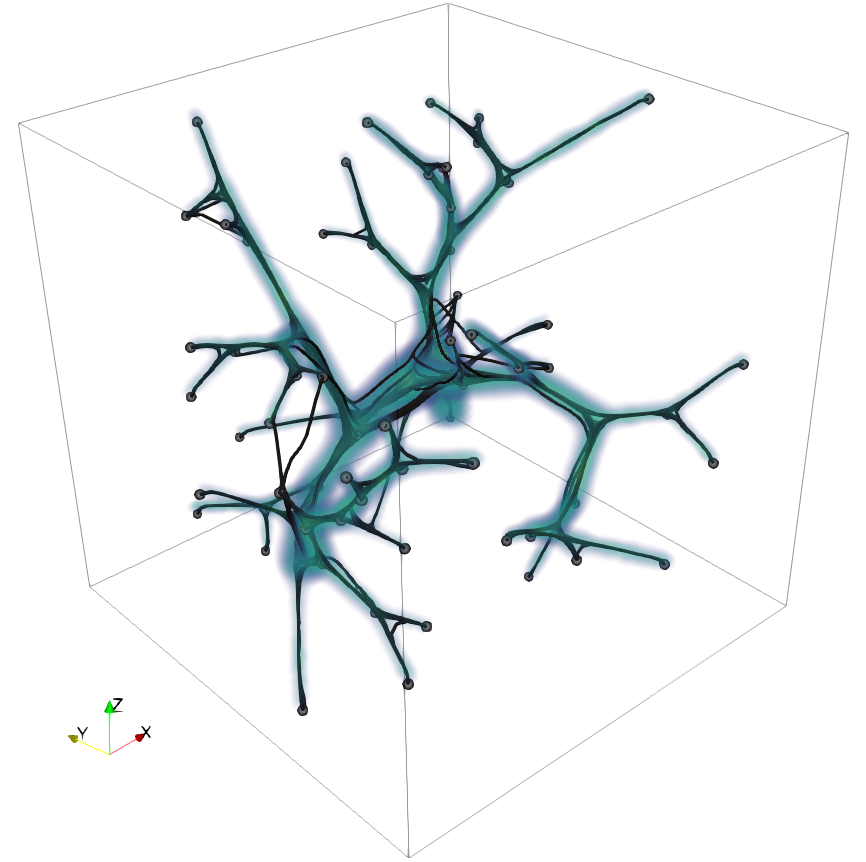}&\includegraphics[width=0.195\textwidth]{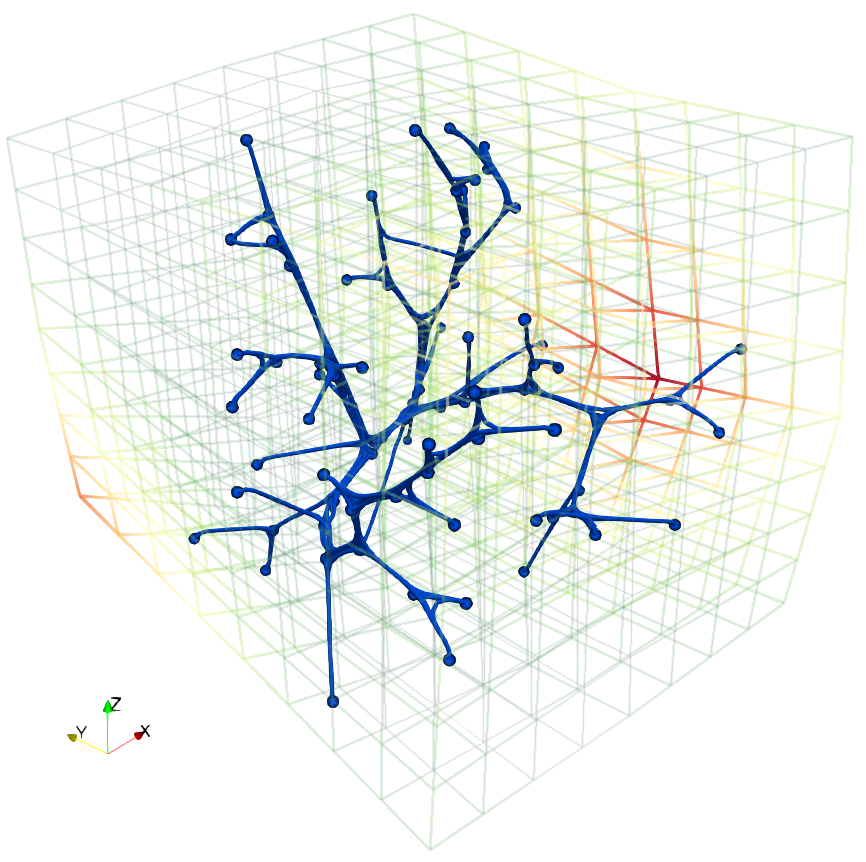}&\includegraphics[width=0.195\textwidth]{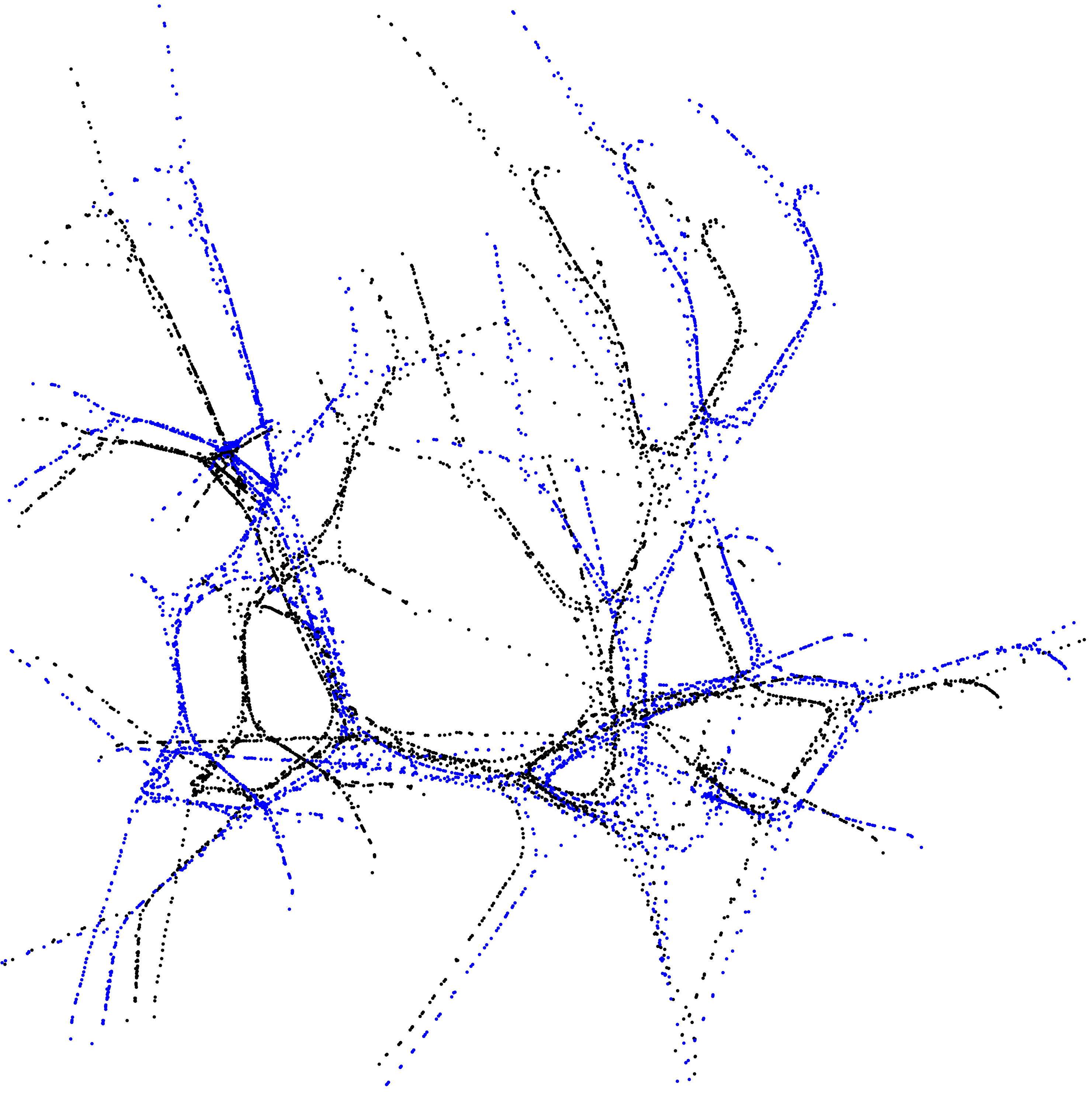}&\multicolumn{2}{c}{\includegraphics[width=0.4\textwidth]{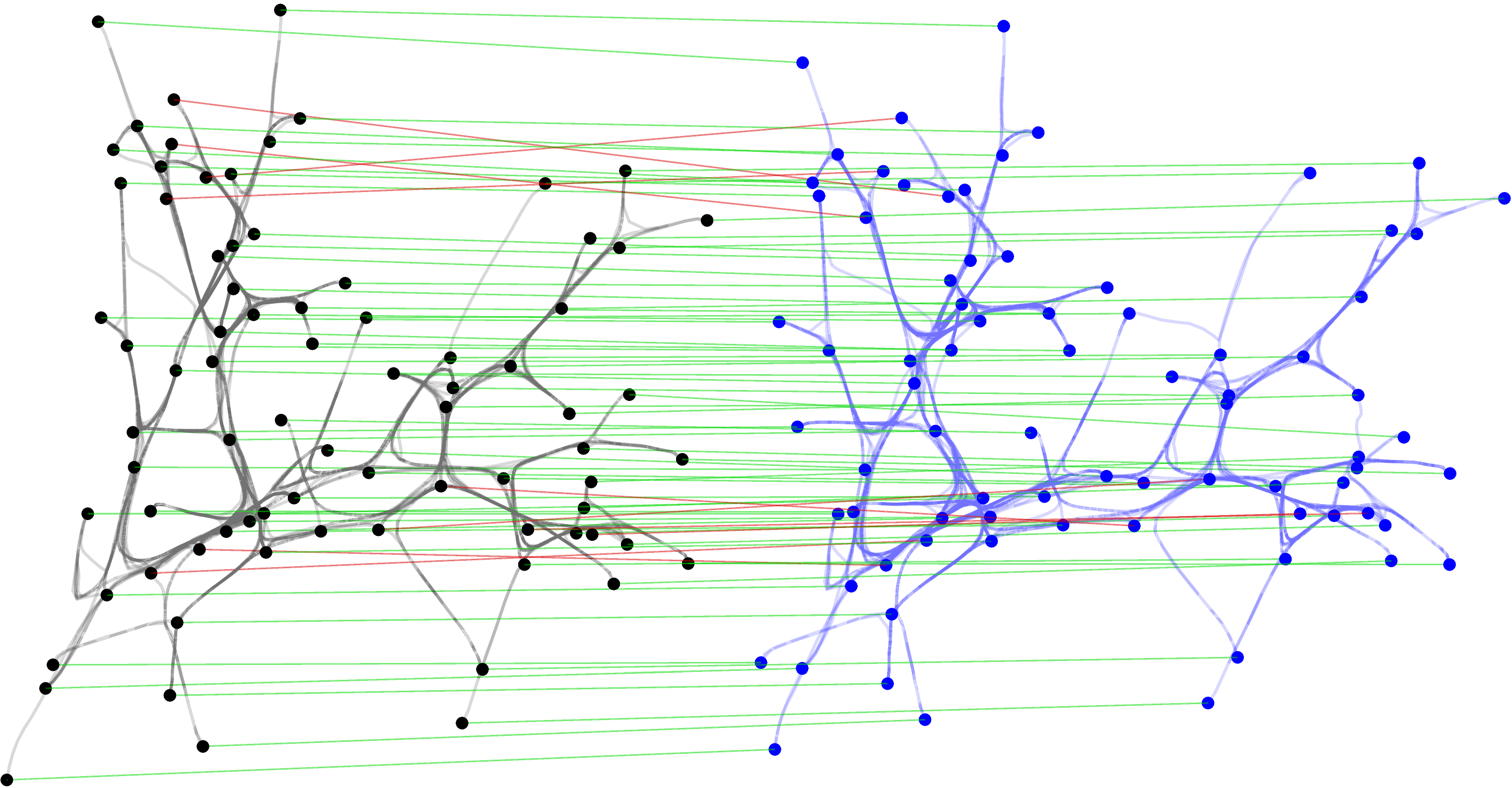}}
\end{tabular}
\normalsize
\vskip -6pt 
\caption{Example of sGVG, simulated deformations, rigid alignment and resulting GM.}
\end{figure}
\paragraph{Dataset.}
A set of 10 synthetic over-connected geodesic vascular graphs (sGVG) and associated minimum spanning trees (sGVT) are obtained from 3D vascular tree images \cite{vascusynth} (isotropic 100$\times$100$\times$100 voxels), as in \cref{subsection_GVG}. Each graph comprises 80 nodes, i.e. the vascular junction and end-points, over-connected within a neighbourhood of radius $\nu = 35$ \mbox{(Fig. 2)}. A total of 10 fully over-connected geodesic vascular graphs (aGVG) as well as the respective minimum spanning trees (aGVT) of the basilar artery are derived as in \cref{subsection_GVG} from Time-of-Flight MRI angiographies (0.35$\times$0.35$\times$0.5 mm), where anatomical vascular junctions and endpoints were manually labelled \mbox{(Fig. 3)} following \cite{jinkins2000atlas}.
\vspace{-6pt}
\paragraph{Synthetic Graphs.}
We randomly deform the synthetic datasets sGVG and sGVT with a non-linear geometrical displacement field (i.e. max magnitude $\mathcal{D}_{30\%}$, $\mathcal{D}_{40\%}$, $\mathcal{D}_{50\%}$ of the graph spatial embedding), a topological pruning (i.e. reducing by $\mathcal{T}_{30\%}$, $\mathcal{T}_{40\%}$, $\mathcal{T}_{50\%}$ the original connectivity), and a combination of both, for a representative set of alterations (Fig. 2). The deformed graphs were then registered with the respective unaltered topologies. The accuracy of the GM is given by the percentage of correct correspondences, and differences of registration performances between sGVG and sGVT are evaluated with a paired Wilcoxon signed rank test.
\vspace{-6pt}
\paragraph{Angiographic Graphs.}
Both aGVG and aGVT are pairwise aligned, covering all possible inter-subject combinations within the same dataset.
The matching accuracy is given by the percentage of correct correspondence among the labelled nodes. Differences between aGVG and aGVT are evaluated with a paired Wilcoxon signed rank test.
\vspace{-8pt}
\subsubsection{Synthetic Graph Matching.}
\label{Res_SynthGM}
In Fig. 4 (charts), the GM accuracy is reported for the synthetic datasets, for each algorithm and for the simulated levels of deformation. The affinity metrics trade-offs are arbitrarily defined as $\mathbf{\alpha} = \left[0.5,0.5\right]$, and $\mathbf{\beta} = \left[0.25,0.25,0.5\right]$ in all cases, to balance the similarity features. Similar trends of performances are observed for the considered GM algorithms across different levels of increasing deformation. Overall, FGM reported the best matching accuracy together with RRWM in both sGVG and sGVT, whereas the other algorithms showed globally varying performances. Purely geometrical displacements did not affect the registration, whereas more severe topological pruning showed a visible drop of accuracy in both sGVG and sGVT, as well as the combination of joint deformations at different degrees. Overall, better matching is found for sGVG compared to sGVT at the same level of  alteration. A significant accuracy drop ($p<0.05$) is found for the registration of tree-like structures, proportional to the combined deformation. This suggests that the proposed registration pipeline would benefit from both geometrical and geodesic information arising from a more dense and redundant over-connected pattern, rather than an explicit vascular tree hierarchy, in presence of non-linear deformations.
\vspace{-8pt}
\subsubsection{Angiographic Graph Matching.}
The accuracy of the pairwise registration for both aGVG and aGVT datasets is reported in Fig. 4 (table). The affinity metrics trade-offs adopted here are the same as those for the synthetic experiments. Overall, discrete matching is obtained for the state-of-the-art FGM ($\text{61.26}\,\pm\,\text{21.91} \%$), as well as for GA ($\text{65.16}\,\pm\,\text{20.39}\%$) and SM ($\text{62.83}\,\pm\,\text{22.96}\%$). The considered angiographic dataset presented large deformations and anatomically different variants (Fig. 3). In line with results of \cref{Res_SynthGM}, the registration of over-connected topologies (aGVG) showed significantly higher accuracy ($p<0.05$), compared to the respective hierarchical minimum spanning trees (aGVT). Globally, nodes mismatch occurred in correspondence of nodes with lower degree and centrality, where higher confusion is found for spatially close vascular end-points and neighbouring branches. Conversely, the correspondence of superior/inferior and left/right branches was correctly preserved in the majority of cases.
\begin{figure}[t]
\label{Fig03}
\tiny
\begin{tabular}{c|c|c}
\textbf{aGVG $\&$ Label set}&\textbf{Rigid Alignment}&\textbf{Fine GM (FGM) - acc: 84.21$\%$}\\
\includegraphics[width=0.249\textwidth]{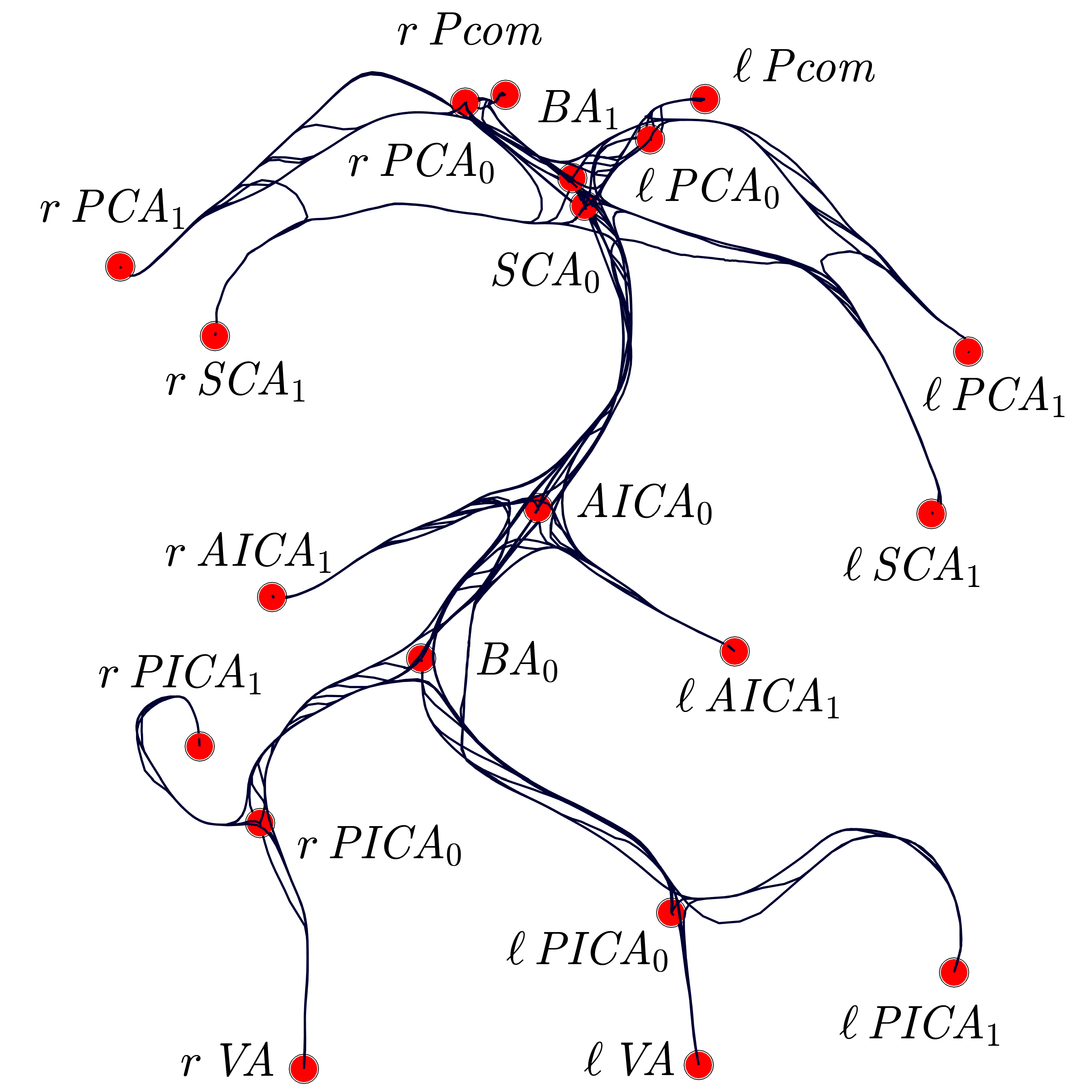}&\includegraphics[width=0.265\textwidth]{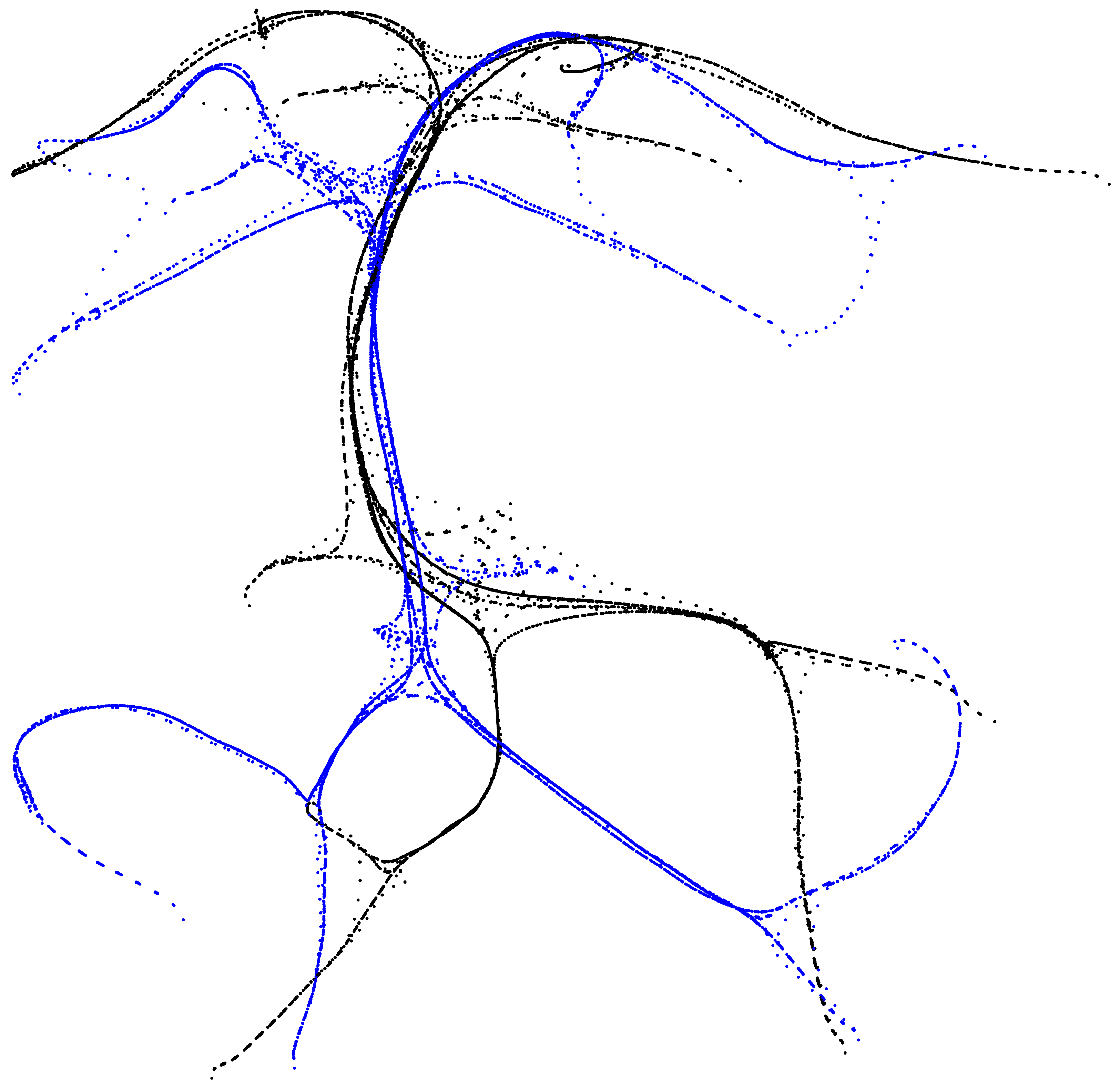}&\includegraphics[width=0.43\textwidth]{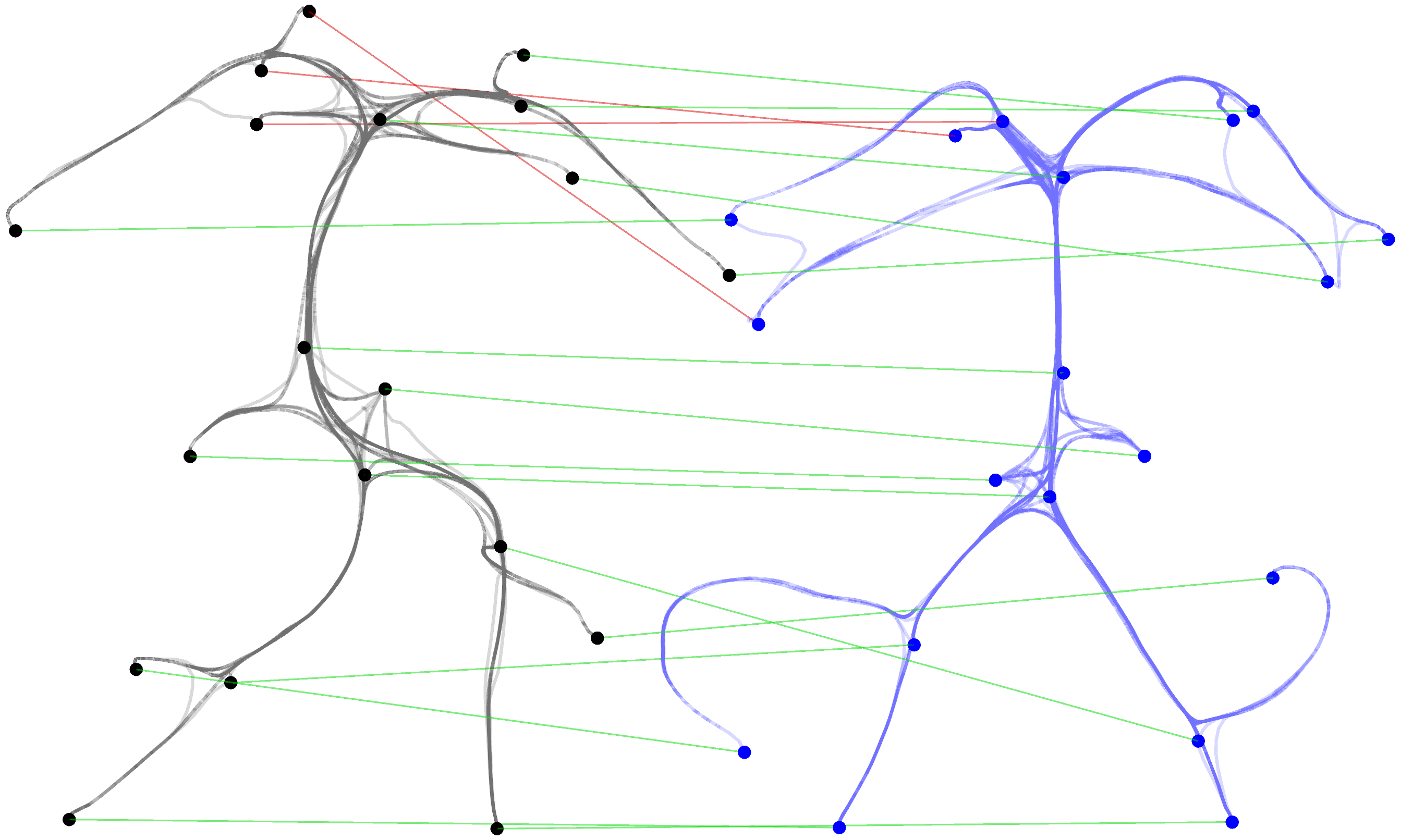}
\end{tabular}
\normalsize
\vskip -6pt 
\caption{aGVG label set \cite{jinkins2000atlas} and pairwise registration of anatomical topologies.}
\end{figure}
\begin{figure}[t]
\label{Fig04}
\tiny
\begin{tabular}{@{}c@{}c|c}
\rotatebox{90}{$~~~~~~~~~~$\textbf{sGVT}[$\%$]$~~~~~~~~~~~~~$\textbf{sGVG}[$\%$]}&\includegraphics[width=0.57\textwidth]{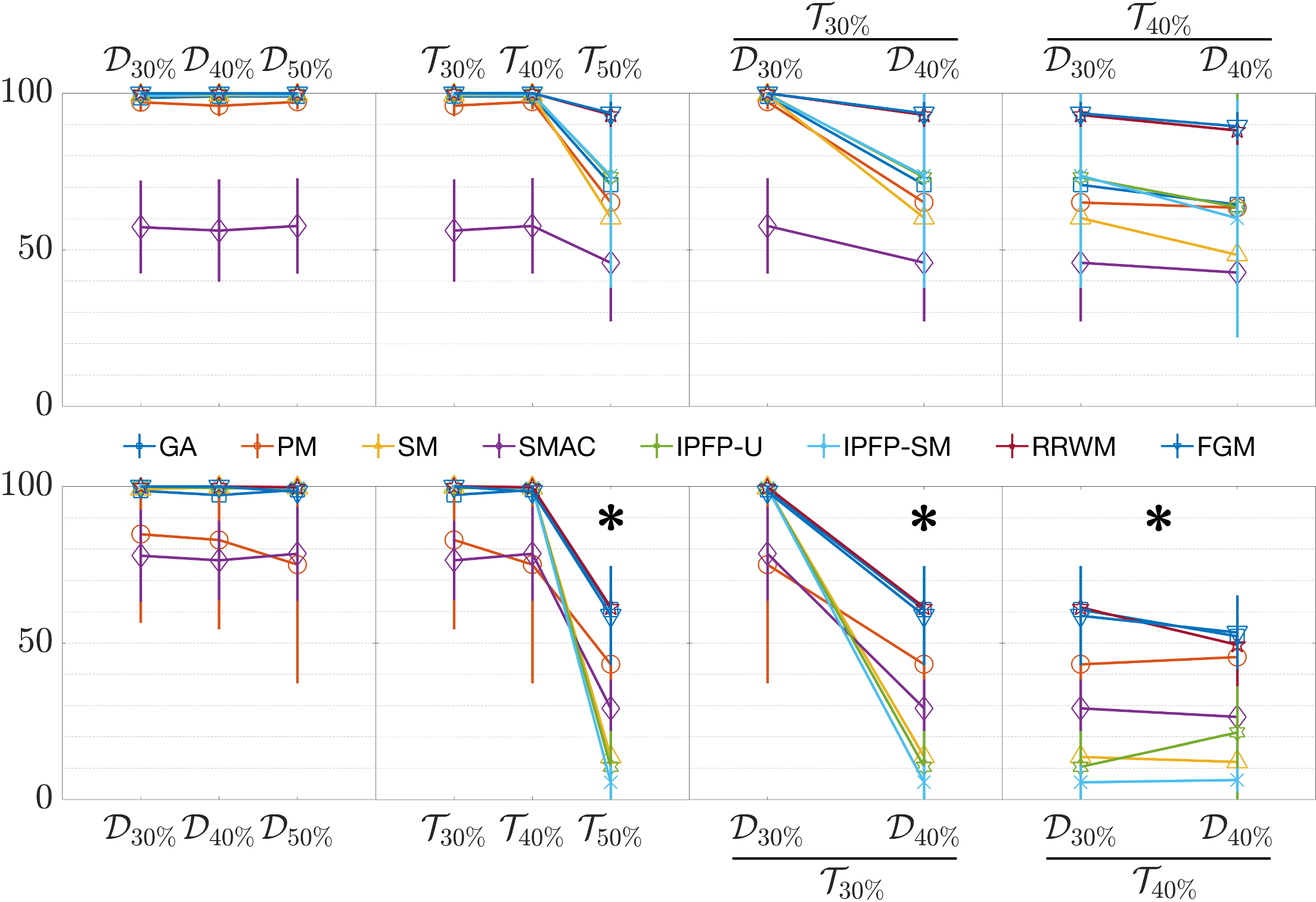}&\raisebox{1.2\height}{
\begin{tabular}{cc|c}
&\textbf{aGVG}[$\%$]&\textbf{aGVT}[$\%$]\\
&&\\
GA&$\,\text{65.16}\,\pm\,\text{20.39}\,$&$\,\text{44.76}\,\pm\,\text{23.37}\,$\\
&($\text{66.67}$)&($\text{44.44}^{*}$)\\
PM&$\,\text{61.72}\,\pm\,\text{23.43}\,$&$\,\text{25.72}\,\pm\,\text{23.91}\,$\\
&($\text{60.86}$)&($\text{20.71}^{*}$)\\
SM&$\,\text{62.83}\,\pm\,\text{22.96}\,$&$\,\text{43.03}\,\pm\,\text{20.38}\,$\\
&($\text{62.07}$)&($\text{42.42}^{*}$)\\
SMAC&$\,\text{41.61}\,\pm\,\text{15.77}\,$&$\,\text{28.63}\,\pm\,\text{18.78}\,$\\
&($\text{40.05}$)&($\text{24.12}^{*}$)\\
IPFP-U&$\,\text{41.59}\,\pm\,\text{16.58}\,$&$\,\text{20.77}\,\pm\,\text{16.19}\,$\\
&($\text{40.59}$)&($\text{16.91}^{*}$)\\
IPFP-SM&$\,\text{38.97}\,\pm\,\text{18.02}\,$&$\,\text{20.96}\,\pm\,\text{13.14}\,$\\
&($\text{37.52}$)&($\text{18.75}^{*}$)\\
RRWM&$\,\text{49.05}\,\pm\,\text{18.31}\,$&$\,\text{44.53}\,\pm\,\text{20.86}\,$\\
&($\text{49.14}$)&($\text{44.12}$)\\
FGM&$\,\text{61.26}\,\pm\,\text{21.91}\,$&$\,\text{48.64}\,\pm\,\text{22.39}\,$\\
&($\text{66.67}$)&($\text{48.28}^{*}$)
\end{tabular}}
\end{tabular}
\vskip -6pt 
\caption{Accuracy of GM: synthetic datasets sGVG vs. sGVT (charts), and angiographic datasets aGVG vs. aGVT (table). Values are mean $\pm$ SD, (median), $~\mathbf{\ast} = p<0.05$.}
\end{figure}
\vspace{-10pt}
\section{Discussion and Conclusions}
\label{section_DiscConc}
\vspace{-6pt}
We presented a vascular graph matching approach to pairwise and elastically register similar topologies, in presence of non-linear deformations. A novel formulation of the vascular network is first introduced using an over-connected geodesic vascular graph. Then, the non-rigid nodes correspondence assignment is solved with a two-steps alignment comprising an optimal rigid registration of the network geometrical embedding, and a set of graph matching algorithms employed in computer vision. For the first time, a general registration of vascular graphs, accounting for noisy over-connected topologies with possible cycles, could be performed by relaxing the explicit hierarchical vessel-tree structure or connectivity patterns specific of a vascular compartment. The use of multiple GM strategies, on the one hand, is motivated by the unconstrained formulation of the GVG, on the other hand, it is justified by the different connectivity lattice of the introduced GVG. The latter can dramatically differ from the connectivity patterns found in computer vision applications (i.e. 3D polygonal subdivision and/or triangulations in 2D), therefore, established GM algorithms may show rather different performances. Early results show, however, good matching from synthetic vascular graphs even in presence of mild-to-moderate non-linear deformations. With the same registration pipeline, we aligned over-connected and redundant topologies, as well as hierarchical undirected tree-structures. Despite these share the same similarity features, the graph matching reported significantly different accuracies, where better nodes correspondences are found for the over-connected topologies. This suggests that the overhead information from the redundant connectivity may enrich the registration space-search with more distinctive cues. Similarly, the registration of geodesic vascular graphs from angiographic datasets reported appreciable matching, even in cases of large spatial deformations and anatomically different topologies, whereas the registration of the associated tree-like structures showed significantly lower accuracies, in line with the synthetic experiments. On the basis of this early evidence, we assume the problem of vascular tree- and graph-registration could be generalised with a multi-spectral network alignment, where further developments towards a more robust design for vascular applications may better incorporate both geometrical and geodesic vascular features.
Although most of the GM algorithms considered in this work are used for 2D applications in computer vision, their general formulation allows the alignment of any generic network, regardless the dimensional embedding, and offer a rich ground for ad-hoc methodological developments.
From a clinical perspective, the successful vascular graph alignment would lead to the definition of a co-registered group-wise prior
to improve the inference of patient specific anatomical topologies.
In last instance, the co-registration of a vascular vectorial prior would pave the way for group-wise analyses with potential applications in neurovascular cross-sectional and longitudinal studies.
\vspace{-6pt}
\subsubsection*{Acknowledgements:}
The study is co-funded from the Wellcome Trust, the EPSRC grant \small EP/H046410/1, \normalsize and the National Institute for Health Research, University College London Hospitals, Biomedical Research Centre.
%
\vspace{-6pt}
\bibliographystyle{abbrv}
\bibliography{./Moriconi2018Elastic_MICCAI18.bbl}
%
%
%
\end{document}